\documentclass[letterpaper, 10 pt, conference]{ieeeconf}  % Comment this line out if you need a4paper

\IEEEoverridecommandlockouts{}                              % This command is only needed if 
\usepackage{graphicx}
\usepackage{amsmath} % assumes amsmath package installed
\usepackage{amssymb}  % assumes amsmath package installed
\usepackage{cite}
\usepackage[bookmarks=true,colorlinks=true,citecolor=magenta,linkcolor=cyan]{hyperref}
\usepackage[ruled,vlined]{algorithm2e}
\usepackage{booktabs} % For formal tables
\usepackage{makecell}
\usepackage{pifont}
\usepackage{float}
\usepackage{subcaption}
\usepackage{array}
\usepackage{siunitx}

\newcommand{\xmark}{\ding{55}}

\usepackage[dvipsnames]{xcolor}

\newcommand{\lars}[1]{{\color{blue}[Lars: #1]}}

\newcommand{\methodname}{JUICER\xspace}
\newcommand{\lamp}{\texttt{lamp}}
\newcommand{\oneleg}{\texttt{one\_leg}}
\newcommand{\squaretable}{\texttt{square\_table}}
\newcommand{\roundtable}{\texttt{round\_table}}
\newcommand{\obschunk}{\mathbf{o_t}}
\newcommand{\actchunk}{\mathbf{a_t}}

\title{\LARGE \bf
JUICER: Data-Efficient Imitation Learning for Robotic Assembly
}

\author{
Lars Ankile$^{1, 3}$, Anthony Simeonov$^{2, 3}$, Idan Shenfeld$^{2, 3}$, Pulkit Agrawal$^{2, 3}$ \\
$^{1}$Harvard University, $^{2}$Massachusetts Institute of Technology, $^{3}$Improbable AI Lab%
\vspace{-5pt}
}

\begin{document}

\maketitle
\thispagestyle{empty}
\pagestyle{empty}

% \iffalse{}
%%%%%%%%%%%%%%%%%%%%%%%%%%%%%%%%%%%%%%%%%%%%%%%%%%%%%%%%%%%%%%%%%%%%%%%%%%%%%%%%
\begin{abstract}
While learning from demonstrations is powerful for acquiring visuomotor policies, high-performance imitation without large demonstration datasets remains challenging for tasks requiring precise, long-horizon manipulation. 
This paper proposes a pipeline for improving imitation learning performance with a small human demonstration budget.
We apply our approach to assembly tasks that require precisely grasping, reorienting, and inserting multiple parts over long horizons and multiple task phases.
Our pipeline combines expressive policy architectures and various techniques for dataset expansion and simulation-based data augmentation. These help expand dataset support and supervise the model with locally corrective actions near bottleneck regions requiring high precision.
We demonstrate our pipeline on four furniture assembly tasks in simulation, enabling a manipulator to assemble up to five parts over nearly 2500 time steps directly from RGB images, outperforming imitation and data augmentation baselines. Project website: \url{https://imitation-juicer.github.io/}.
\end{abstract}
\IEEEpeerreviewmaketitle

\vspace{-5pt}
%%%%%%%%%%%%%%%%%%%%%%%%%%%%%%%%%%%%%%%%%%%%%%%%%%%%%%%%%%%%%%%%%%%%%%%%%%%%%%%%
\section{Introduction}

Automatic assembly is one of the most practically valuable applications of robotic technology.
However, robots have demonstrated limited utility in production environments where a wide variety of products are manufactured in small quantities (i.e., ``high-mix, low-volume'').
% However, robots have demonstrated limited utility in high-mix, low-volume production\footnote{Production environments where a wide variety of products (high-mix) are manufactured in small quantities (low-volume).} operations that require significant adaptability and frequent retooling.
%
Robot learning has the potential to transform this landscape. For instance, a programmer can directly ``teach'' new assembly sequences via demonstration rather than scripting exact trajectories.
Learned controllers that work directly with sensor data can also reduce reliance on custom part-positioning fixtures.
However, learning to perform assembly from raw perception has remained challenging~\cite{heo_furniturebench_2023}, as assembly requires long-horizon manipulation with high precision requirements.
%
% This paper aims to address some of these performance gaps by suggesting techniques that improve Imitation Learning (IL) from a modest number (50) of demonstrations. % in simulation.
%

%
To train closed-loop assembly policies that use RGB images, one could try using Reinforcement Learning (RL), but RL struggles with long task horizons and sparse rewards. 
On the other hand, while learning from demonstrations is more tractable, requiring many demonstrations creates a significant data collection burden.
Instead, we consider Imitation Learning (IL) in a small-data regime that enables users to collect demonstration data themselves ($\sim$50 demonstrations).
One could also combine IL with RL~\cite{Rajeswaran-RSS-18_dapg} or fine-tune a pre-trained multitask model~\cite{octo_2023}, but those investigations introduce additional complexity and are beyond the scope of this work. 
Thus, the primary challenge is to extract as much information and performance as possible from a modestly sized, manually collected dataset of assembly trajectories.
%
% We wish to understand the resulting challenges and suggest a pipeline that improves IL under these conditions. 
Our goal is to understand the resulting challenges that emerge and suggest a pipeline that improves Imitation Learning when operating with a modest number of demonstrations. % in simulation.

For one,
%even with a modestly sized dataset, 
effectively fitting a complex set of demonstrated actions while operating from raw images can be challenging for long-horizon tasks requiring high precision.
The choice of policy architecture and action prediction mechanism has a tremendous impact on the ability to fit the data well.
Building off recent work, we provide additional evidence for the advantages of representing policies as conditional diffusion models~\cite{janner_planning_2022,ajay_compositional_2023,chi_diffusion_2023,pearce_imitating_2023} and predicting chunks of multiple future actions~\cite{chi_diffusion_2023, zhao_rss23_aloha}.

Another difficulty is learning robust behaviors around ``bottleneck'' regions where even slight imprecisions can lead to failure. For example, mistakes during part insertion frequently lead to dropped parts and overall task failure.
%
% Structured data augmentation and noising can help mitigate this by expanding dataset support.
One strategy to mitigate this is to expand dataset support via structured data augmentation and noising.
The main idea is to supervise the model with locally corrective actions that return to the training distribution from synthetically perturbed states~\cite{ke_grasping_2021,ke_ccil_2023,block2024provable,zhou2023nerf}.
However, such methods often make assumptions that prevent their direct application to visuomotor policy learning for multi-step assembly (e.g., known object poses, static scenes, assuming no object in hand).
We deploy a similar strategy that requires a weaker assumption - the ability to reset the system to a given state in the demonstrations. % (we currently use simulation to enable such resets).
%in simulation that alleviates several of their assumptions.
Our current work uses a simulator to perform such resets (and also evaluates learned policies in simulation), but automatic resets can be deployed in real-world settings as well~\cite{gupta2021reset}.
We reset the scene to bottleneck states, perturb the scene by simulating random ``disassembly'' actions, and synthesize corrective actions while rendering new scene images.
The corrective actions are obtained by reversing the disassembly sequence~\cite{spector_insertionnet_2021,spector_insertionnet_2022, zakka2020form2fit,tian2022assemble_them_all}.
This enables structured data noising in a broader class of scenarios. %, with the tradeoff of requiring bottleneck state annotations and simulation for resets and rendering. 

\begin{figure*}[t]
    \centering
    \includegraphics[width=0.95\textwidth]{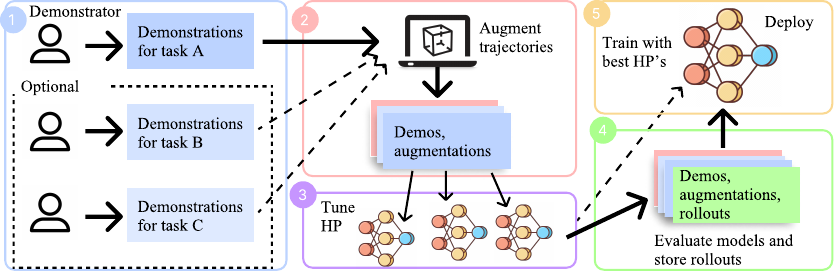}
    \caption{\small Overview of the proposed approach. (1) Collect a small number of demonstrations for the task (and related tasks, if available). (2) Using annotations of \textit{bottleneck states}, augment the demonstration trajectories to create synthetic corrective actions and increase coverage around the bottleneck states. (3) Use this dataset to train models with different hyperparameters. (4) Store all rollouts throughout model evaluations. (5) Add any successful rollout to the training set and train the best architecture on all data amassed.}
    \label{fig:system-overview}
    \vspace{-18pt}
\end{figure*}

Finally, we consider opportunities to automatically expand the dataset of whole trajectories.
We take advantage of the iterative model development cycle across tasks to achieve this. Namely, successful or partially successful rollouts are often collected during model evaluation if the policy has a non-zero success rate.
New data may also be collected (either via demos or evaluation rollouts) for other tasks introduced in parallel. We use these in an IL-based ``collect and infer'' setup~\cite{riedmiller2022collect_infer,lampe2023mastering_stacking,bousmalis2023robocat} and obtain larger, more diverse datasets without incurring additional human effort.

In summary, we propose \methodname, a pipeline for learning high-precision manipulation from a small number of demonstrations. Our pipeline combines the benefits of diffusion policy architectures and mechanisms for automatically expanding dataset size via data noising and iterative model development cycles. 
We show the technique's effectiveness on four simulated tasks from the FurnitureBench benchmark~\cite{heo_furniturebench_2023} for long-horizon furniture assembly, including tasks with horizons up to $\sim$2500 timesteps and assemblies of up to 5 parts to be precisely grasped, oriented, and inserted. 
Our results show that each component of our system improves overall task success compared to vanilla IL baselines. Our ablations illustrate how different design choices impact overall system performance, and we demonstrate that \methodname enables learning high-performance policies using as few as 10 human demonstrations.
Finally, we contribute our bottleneck state labeling tool, trajectory augmentation tool, collected demonstration datasets, and IL-based collect-and-infer pipeline for the community to utilize and build on.
%

% \vspace{-1pt}
\section{Preliminaries}
% \vspace{-2pt}

\subsection{Notation and System Components}

We aim to learn a policy $\pi_\theta(a_t|o_t)$ mapping observations $o_{t}$ to actions $a_{t}$ using a dataset of trajectories, $\mathcal D=\{\tau_1, ..., \tau_N\}$, $\tau_i=\{(o_1, a_1), ..., (o_T, a_T)\}$, with $T$ being the trajectory length.
Our observations include RGB images (front and wrist views) and the robot's proprioceptive state. The proprioceptive state contains the current end-effector pose and velocity, and current gripper width. Our actions represent desired 6-Degrees-of-Freedom (DoF) end-effector poses, reached via a low-level Operation Space Controller~\cite{khatib1987unified_osc}, and a gripper open/close action.
We train $\pi_{\theta}$ with Behavior Cloning (BC), i.e., we optimize $\theta$ to maximize the likelihood of the data, $\underset{\theta}{\text{argmax}}\ \mathbb E_{(a_t,o_t)\sim\mathcal D} \left[\log\pi_\theta(a_t|o_t) \right]$. 

The policy input and output can consist of consecutive observation or action chunks.
Denoting $T_o$ as the number of observations we condition on (i.e., up to and including the current step), we pass $\obschunk = [o_{t-T_o}, ..., o_t]$ to $\pi_{\theta}$, with each $o_{t}$ of dimension $|o|$.
Let $T_p$ be the number of predicted future actions, i.e., the policy outputs a sequence of actions $\actchunk = [a_t, ..., a_{t+T_p}]$, with each $a_{t}$ of dimension $|a|$. Although predicting an action chunk $\actchunk$ of length $T_p$, we only execute a subset $[a_{t},...,a_{T_a}]$, with execution horizon $T_a \leq T_p$.

\subsection{Denoising Diffusion Probabilistic Models}

Denoising Diffusion Probabilistic Models (DDPM) \cite{ho_denoising_2020,song2020score,sohl2015deep_diffusion} are a class of generative models used to learn high-dimensional distributions. DDPMs work by gradually adding noise to data and then learning to reverse this process.

The model consists of two main processes:
(1) A fixed forward process that progressively adds Gaussian noise to the data.
(2) A learned reverse process that removes noise step-by-step to generate new samples.

During training, the model learns to predict and remove noise from partially noised inputs using a noise prediction loss.

At inference time, generation starts from pure Gaussian noise. The model then iteratively refines this noise into a coherent sample by repeatedly predicting and subtracting noise. Techniques like the DDIM sampler \cite{song_denoising_2022,lu2022dpm} allows for faster, deterministic sampling with fewer steps.

For a detailed mathematical treatment of DDPMs, including the specific equations for the forward and reverse processes, please refer to \autoref{app:denoising-diffusion}.

\section{Methods}

\subsection{System Overview}

We propose JUICER, a data-efficient Imitation Learning (IL) pipeline for precise and long-horizon assembly from image observations in the simulator, presented in \autoref{fig:system-overview}. The pipeline starts by \textbf{(1)} collecting a small number of task demonstrations (\autoref{sec:data-collection}) (as well as any available related tasks). The demonstrator also annotates states that require extra precision. These labels are used to \textbf{(2)} synthetically increase the data support around these states (\autoref{sec:trajectory-augmentation}). To model diverse human-collected and synthetic data, we \textbf{(3)} fit a Diffusion Policy (\autoref{sec:model-architecture}). As models are evaluated, any (partially) successful trajectory is \textbf{(4)} stored and added back to the training set (\autoref{sec:collect-and-infer}), further increasing support. Using the best hyperparameters and all data amassed in the process, we \textbf{(5)} train a final model for initial deployment.

\subsection{Data Collection and Annotation}
\label{sec:data-collection}

The first step is robot teleoperation. While any teleoperation interface can be used, our system utilizes a 3DConnexion SpaceMouse for 6-DoF end-effector control. 
For each task, 50 demos are collected, taking 1-3 hours per task. 
Teleoperation in simulation offers unique opportunities to make demo collection more seamless and improve user experience. 
For instance, using the ability to perform resets, we implement an ``undo'' button that restores the state 10 frames earlier, enabling users to correct mistakes instantly. 
These additions save time by reducing the number of trials discarded due to operator failure.

After data collection, we add sparse annotations of ``bottleneck'' state transitions requiring high precision (these labels are used for data augmentation, see \autoref{sec:trajectory-augmentation}).
To annotate the demonstrations, a user loads the successful trajectories, steps through each episode, and marks the timesteps when bottleneck states occur.
This step adds $\sim$15\% to the total data collection time, and even the longer demos ($\sim$2500 time steps) can be fully annotated in $\sim$1-2 minutes.
We perform the annotation in a separate step as we found it hard for the demonstrator to remember to label bottleneck states accurately while providing demos, thus reducing the cognitive load.

\subsection{Trajectory Augmentation}
\label{sec:trajectory-augmentation}

\begin{figure}
    \centering
    \includegraphics[width=0.50\textwidth]{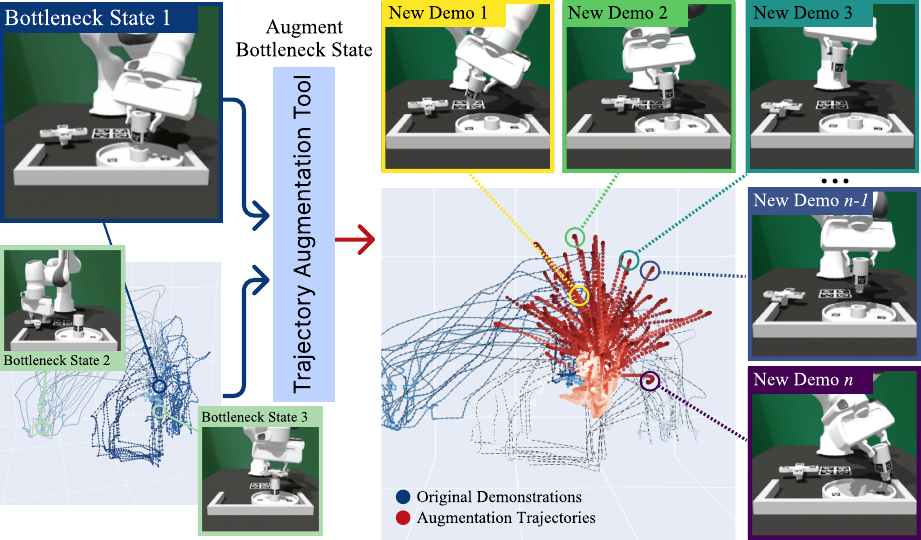}
    \caption{\small An example of extracting \textit{``bottleneck''} states and using trajectory augmentation tool to create an arbitrary amount of counterfactual trajectories near the bottleneck, effectively increasing the data support and teaching corrective actions.}
    \label{fig:trajectory-augmentation}
    \vspace{-16pt}
\end{figure}

Our experiments showed that small misalignment errors during precise grasps or insertions could quickly compound and lead to catastrophic task failure (e.g., parts falling out of reach in extreme out-of-distribution poses).
We, therefore, want to improve the model's ability to reach and transition through these bottleneck regions robustly.
%
% However, we also want to avoid requiring the demonstrator to provide examples of additional recovery behaviors manually~\cite{ross2011_dagger}. 
However, we want to avoid having the demonstrator manually provide examples of additional recovery behaviors~\cite{ross2011_dagger, brandfonbrener_visual_2022}. 
Our solution is to increase dataset support in regions that require high precision via synthetic data augmentation (see \autoref{alg:trajectory-augmentation}).
The first step is to reset the simulated robot and environment to a recorded bottleneck state from a demonstration. We then sample a random end-effector pose \emph{away} from the bottleneck state and construct a sequence of ``disassembly'' actions that reach this pose via (spherical) linear interpolation.
Finally, starting from the random pose, the robot executes a ``correction'' that returns from the randomly perturbed state to the original bottleneck. This correction is obtained by \emph{reversing} the disassembly actions. 
We record the corrective actions and rendered scene images and compare the resulting scene state to the original demonstration.
If the final states match (within an error $\epsilon$), the trajectory segment is saved in the same format as the demonstrations to be used for BC training. Some sampled trajectories are infeasible causing other parts to move in the backward portion, necessitating the need for success check after the forward portion. In our experiments, augmentation transitions will constitute $\sim$10-20\% of the total number of transitions (see \autoref{app:dataset-sizes} for exact data mixture numbers).

\autoref{fig:trajectory-augmentation} shows an example of this procedure for the \roundtable{} task. Note how the augmented trajectory dataset provides a ``funnel'' toward the critical state required for successful insertion. 
This blends prior techniques for guiding tasks like assembly~\cite{tian2022assemble_them_all}, insertion~\cite{spector2021insertionnet1,spector2022insertionnet2}, and kitting~\cite{zakka2020form2fit}, together with structured data noising that has been deployed to reduce covariate shift in IL~\cite{ke_grasping_2021,zhou2023nerf}.

% \newlength{\textfloatsepsave}
% \setlength{\textfloatsepsave}{\textfloatsep}
\setlength{\textfloatsep}{5pt}
% \begin{figure*}[t]
\begin{algorithm}[t]
\DontPrintSemicolon
\SetAlgoLined
\KwIn{Spherical coordinate limits, $\mathbf{r}=\{r_{\text{min}}, r_{\text{max}}\}$, $\boldsymbol{\theta}=\{\theta_{\text{min}}, \theta_{\text{max}}\}$, $\boldsymbol{\phi}=\{\phi_{\text{min}}, \phi_{\text{max}}\}$;\\
Dataset of human-collected demos $\mathcal{D}_H$}
\KwResult{Augmented trajectories $\mathcal{D}_A=\{\tau_i\}_{i\in [N_A]}$}\;
\textbf{Init:} $i \leftarrow 1$, $\mathcal{D}_A=\varnothing$ \;
\While{$i < N_A$ \KwTo $N_p$}{
  Sample demonstration trajectory $\tau_H\sim \mathcal{D}_H$ \;
  Sample bottleneck state $s_{\text{bottleneck}}\sim\tau_H$
  
  Reset world state to $s_{\text{bottleneck}}$\;
  Sample a state $s_{target}$ within the limits $\mathbf{r},\boldsymbol{\theta},\boldsymbol{\phi}$ \;
  Command EE from $s_{\text{bottleneck}}$ to $s_{target}$ and record inverse actions $\tau_i^\text{cand}$ \;
  Execute $\tau_i^\text{cand}$ from $s_{target}$, ending in state $s^\prime$ \;
\If{$\parallel s^\prime - s_{\text{bottleneck}}\parallel^2 \leq \epsilon$}{
    Store trajectory $\mathcal D_A \leftarrow \mathcal D_A \cup \{\tau_i^\text{cand}\}$ \;
    $i \leftarrow i + 1$
}
\Else{
    Discard candidate trajectory $\tau_i^\text{cand}$
}
}
\Return{$\mathcal D_A$}
\caption{Backward Trajectory Augmentation}
\label{alg:trajectory-augmentation}
\end{algorithm}
% \vspace{-6pt}
% \end{figure*}
% \setlength{\textfloatsep}{\textfloatsepsave}

\subsection{Policy Design Choices}
\label{sec:model-architecture}

\subsubsection{Data Processing and Image Encoding}
We normalize all actions and proprioceptive states so each dimension lies in the range $[-1, 1]$ using $\min$ and $\max$ statistics from demo data across all tasks.
Consistent with prior work, we found position control to outperform delta control\cite{chi_diffusion_2023}. In position control, the action directly specifies the target position of the end-effector, while in delta control, the action represents the change in position relative to the current state. We represent orientations using a 6-dimensional representation shown to be easier to learn \cite{zhou2019continuity,levinson2020analysis}, resulting in an action space of $|a_t|=10$ (see details in \autoref{app:rotation-representation}).
When processing image observations, we first resize raw RGB images from $1280\times 720$ to $320\times 240$. We then perform random cropping and color jitter to increase dataset diversity and reduce overfitting, and perform a final resizing to $224\times 224$.

We encode the images with a vision model. The encoder outputs latent image representations that we project to 128-dimensional vectors (one for each camera view). The final observation provided to the policy is of size $|o_{t}|=2\cdot 128 + 16 = 272$\footnote{16 dimensions comes from 3 position coordinates, 6 for orientation, 3 for linear velocity, 3 for angular velocity, and 1 for gripper action.}.
%
% Contrary to~\cite{chi_diffusion_2023}, we find a standard \texttt{ResNet18} pre-trained on ImageNet to outperform training a \texttt{ResNet18} with \texttt{SpatialSoftmax} pooling, similar to\cite{pearce_imitating_2023}.
Contrary to~\cite{chi_diffusion_2023}, we find an ImageNet-pre-trained \texttt{ResNet18} with global pooling to outperform a \texttt{ResNet18} trained from scratch with \texttt{SpatialSoftmax} pooling, similar to\cite{pearce_imitating_2023}.
Each \texttt{ResNet18} has $\sim$11M parameters.

\subsubsection{Diffusion Policy}

% Given a dataset of state-action chunks generated by the data distribution $q(\mathbf o, \mathbf a)$, $\mathcal{D}=\{(\mathbf o_i, \mathbf a_i)\}_{i\in[N]}$, the goal is to fit a model to estimate a policy $\pi_\theta(\mathbf a_t|\mathbf o_t)$.
Given a dataset of observation-action chunks, $\mathcal{D}=\{(\mathbf o_i, \mathbf a_i)\}_{i\in[N]}$, the goal is to fit a policy model $\pi_\theta(\mathbf a_t|\mathbf o_t)$.
The standard denoising objective in fitting diffusion models from \cite{ho_denoising_2020} is adapted to the setting of modeling an action distribution conditioned on an observation, $p(\mathbf{a}_t|\mathbf{o}_t)$, but where the noise model is parameterized as $\epsilon_\theta(\mathbf{a}_t,\mathbf{o}_t,k)$.
Both the action and observation are chunks along the time-dimension, $\mathbf{a}_t\in\mathbb R^{T_p\times |a_{t}|}$ and $\mathbf{o}_t\in\mathbb R^{T_o\times |o_{t}|}$.
We find longer prediction horizons benefit performance and use $T_p=32$ in all our experiments. We still find the action execution horizon of $T_a=8$ provides a good tradeoff between open-loop consistency and closed-loop reactivity. Consistent with recent work\cite{bousmalis2023robocat}, we have not found including history helpful and use $T_o=1$ for all tasks. Since velocity is included in the proprioception, multiple frames is not needed to infer it.

Like prior work\cite{ajay_is_2023,chi_diffusion_2023}, we use a 1D temporal CNN-based U-Net introduced by\cite{janner_planning_2022}, with the same FiLM conditioning layers in the residual blocks used in~\cite{chi_diffusion_2023}.  
Experiments with the time-series diffusion transformer introduced in\cite{chi_diffusion_2023} did not improve performance in our tasks.
Unless stated otherwise, our experiments use a U-Net with 3 downsampling blocks of dimension 256, 512, and 1024, yielding a model with 69 million parameters and a full policy model with 91 million parameters.
At training time, we used 100 denoising steps for the DDPM sampler\cite{ho_denoising_2020} and 8 denoising steps with the DDIM sampler\cite{song_denoising_2022,lu2022dpm} at inference time.
Hyperparameters were chosen by starting with best-practices presented in prior work as the nominal and then sweeping over different parameter choices. The full set of hyperparameters used is presented in \autoref{app:hyperparameters}.

\begin{figure*}[t]
    \centering
    \includegraphics[width=0.95\textwidth]{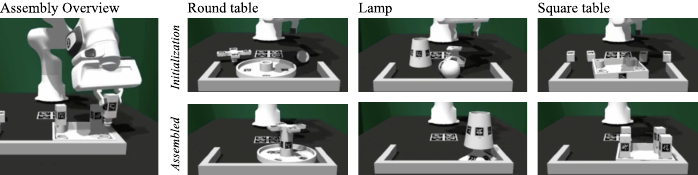}
    \caption{\small Overview of the tasks. The first row shows a random initialization of the parts, and the second row shows the full assembly.}
    \label{fig:task-overview}
    \vspace{-18pt}
\end{figure*}

\subsection{Dataset Expansion with ``Collect-and-Infer''}
\label{sec:collect-and-infer}
After training on the combined teleop and augmentation trajectories, we can deploy initial versions of the assembly controller and evaluate their performance. Through this process, we expand the dataset by re-incorporating any successful rollouts into the training set and training a new model on the expanded dataset. Concretely, starting with the initial dataset $\mathcal{D}_H$ (of 50 demos), we fit $k$ policies $\pi_{\mathcal{D}_{H}}^i, i\in[k]$ with different hyperparameters. We then evaluate these policies and store any successful trajectories in $\mathcal{D}_E = \{\tau_i^j\}, i\in[k], j\in[n_\text{eval}]$. Finally, we take the best policy and train a new one from scratch with the same hyperparameters on all the data, $\mathcal{D}_{H+E} = \mathcal{D}_{H} \cup \mathcal{D}_{E}$. This paper focuses on the initial effects of the technique and limits the rollouts used to 50 (experiments with more than 50 rollouts are included in the Appendix). % \lars{Mention something like this? Also, mention that experiments with more than 50 are in the supplementary?}.

This is reminiscent of the ``collect-and-infer'' paradigm for RL~\cite{riedmiller_collect_2021,lampe_mastering_2023} and IL~\cite{bousmalis2023robocat} which has shown to be a simple but powerful mechanism for iterative policy improvement. We find that collect-and-infer complements the augmentation procedure discussed \autoref{sec:trajectory-augmentation}. Namely, trajectory augmentation is directly applied to original demonstrations and does not require collecting new rollouts. We find, though, that performance improvements saturate unless more full trajectory data is available. 
On the other hand, while collect-and-infer requires an initial policy with non-zero success, it provides valuable extra supervision of \emph{complete} trajectories. Our experimental results show that combining both dataset expansion methods boosts overall performance the most. 

\vspace{-4pt}
\subsection{Multitask Learning}
\label{sec:multitask-learning}
Prior work has shown positive transfer between different tasks on both large~\cite{brohan_rt-2_2023} and more modest scales~\cite{bousmalis2023robocat}.
Similarly, our final step involves co-training on data from other related assembly tasks.
This is inspired by observations in other prior works that show that multitask training benefits performance even at a modest dataset scale (100-1000 demos)~\cite{bousmalis2023robocat,shridhar2022peract}.

\section{Experiments and Results}
\label{sec:experiments}

\subsection{Experimental Setup}

\subsubsection{Tasks}
We choose 4 tasks from the benchmark FurnitureBench\cite{heo_furniturebench_2023}: \roundtable{}, \lamp{}, \squaretable{}, and \oneleg{} (where \oneleg{} is a subtask of the \squaretable{}, see \autoref{fig:task-overview}). We note that while the environment includes fiducial markers on the parts, our approach does not utilize these markers at all, relying solely on raw RGB images. This task subset presents a variety of challenges which are summarized in \autoref{tab:task-attributes}.

\begin{table}[htbp]
\vspace{-8pt}
\centering
\caption{Task Attribute Overview}
\label{tab:task-attributes}
\begin{tabular}{@{}lcccc@{}}
\toprule
                       & \makecell{Round Table} & Lamp       & \makecell{Square Table} & \makecell{One Leg} \\ \midrule
Episode Length         & 1,100                     & 900        & 2,500                      & 550                   \\
\# Parts to Assemble   & 3                         & 3          & 5                          & 2                     \\
Rolling Object       & \xmark                    & \checkmark & \xmark                     & \xmark                \\
\# Precise Insertions  & 2                         & 1          & 4                          & 1                     \\
Precise Grasping       & \checkmark                & \xmark     & \xmark                     & \xmark                \\
Insertion Occlusion    & \checkmark                & \xmark     & \xmark                     & \xmark                \\ \bottomrule
\end{tabular}
\vspace{-6pt}
\end{table}

\subsubsection{Training}

\paragraph{Default Training protocol}

For each task and dataset, we train 5 models (as described in \autoref{sec:model-architecture}) from different seeds for up to 160,000 gradient steps with the \texttt{AdamW} 
optimizer\cite{loshchilov2017decoupled}. We use a learning rate of $10^{-4}$ and a cosine learning rate schedule\cite{loshchilov2016sgdr} with 500 warmup steps and a batch size of 256. To avoid overfitting, we terminate if validation loss does not improve for 5 epochs.

During training, we apply image augmentations to both the wrist and front camera images, similar to\cite{laskin_reinforcement_2020,lee2021beyond}. We apply the PyTorch color jitter augmentation with the parameters for brightness, contrast, saturation, and hue, all equal to $0.3$, and Gaussian blur with kernel size 5 and $\sigma\in [0.01, 2]$ to both views.
We also apply random image cropping, but only on the front view. We do not crop the wrist view because the end-effector action labels are not invariant to randomly translated crops of the wrist view. 

\paragraph{Multitask Training}

To investigate the impact of multitask co-training, we train models on mixes of 50 demonstrations for each of the \roundtable{}, \squaretable{}, and \lamp{} tasks, for a total of 150 demos.

\subsubsection{Evaluation} We evaluate the 5 models trained from different random initializations per condition on 100 rollouts and report the mean and max \textit{success rate}, i.e., the share of attempts that result in a complete assembly. This is measured with comparing relative part poses to the desired pose given by the CAD model in the fully assembled configuration.
We also examine what performance can be achieved using an even smaller budget of human-collected demos (see \autoref{sec:minimize-demo-budget}).

\subsubsection{Baselines}

The baseline architecture is a residual MLP model with 5 residual blocks ($\sim$10 million parameters). It receives the same features as the diffusion model, i.e., image encoding vectors and the proprioceptive state.
The first baseline is an MLP without action-chunking, i.e., $T_p=T_a=1$. The second uses the same action-chunking as the diffusion policy, $T_p=32$ and $T_a=8$.
We also compare against the noising approach proposed in~\cite{ke_grasping_2021}, which suggests expressing robot states in an object-centric coordinate frame and perturbing robot states to synthesize corrective action labels and reduce covariate shift. Under our setting, where the policy cannot access object poses, this reduces to injecting noise \emph{only} in the robot's proprioceptive state and leaving the images unmodified.

% \iffalse{}
\iftrue{}
% \begin{figure*}[htbp]
\begin{figure*}[t]
    \centering
    \begin{minipage}{0.6\textwidth}
        \centering
        \vspace{3pt}
        \includegraphics[width=1.0\linewidth]{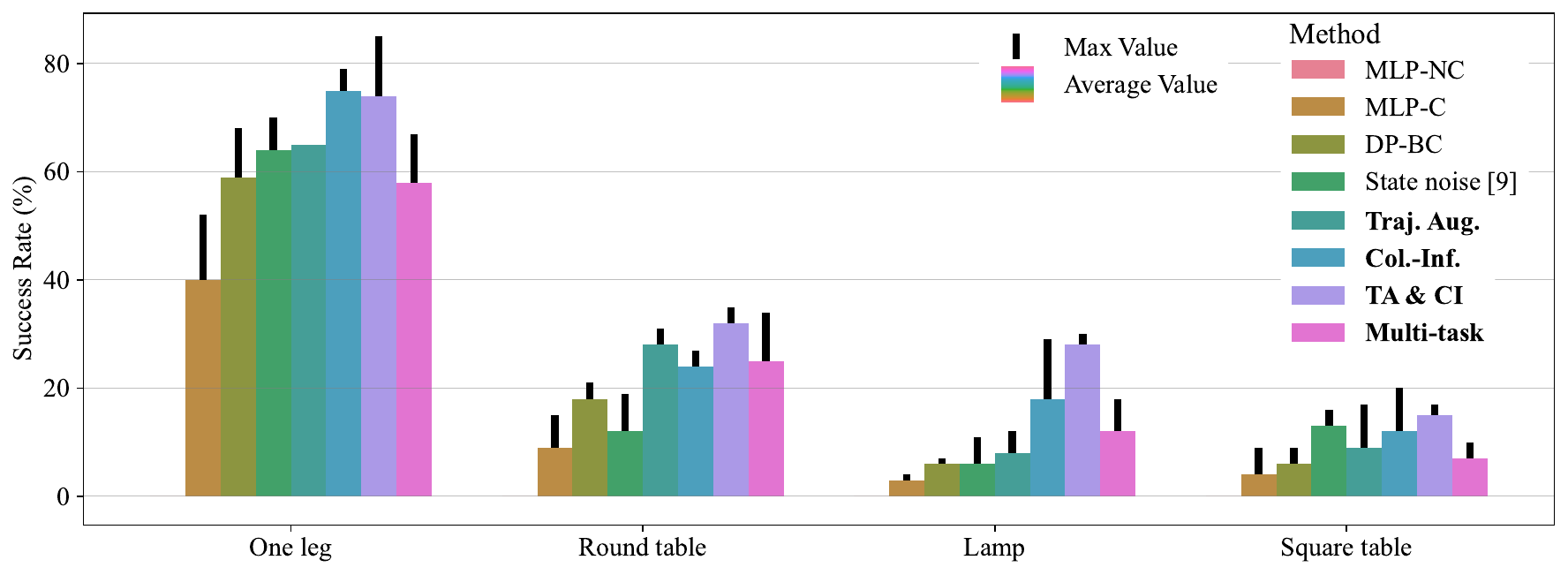}
        % \vspace{1pt}
        \captionof{figure}{\small Average and maximum success rates (\%) of methods across tasks. Bolded methods are components of our \methodname pipeline.} % \pulkit{Good to identify which methods are our contribution. Also, seems like the error bars are above the mean for all bars -- seems a bit unintuitive.}} %Colored bars and black upward bars denote average and maximum success across five training runs.}
        \label{fig:main-results}
    \end{minipage}%
    \hfill
    \begin{minipage}{0.365\textwidth}
        \centering
        \vspace{-19pt}
        \includegraphics[width=0.87\linewidth]{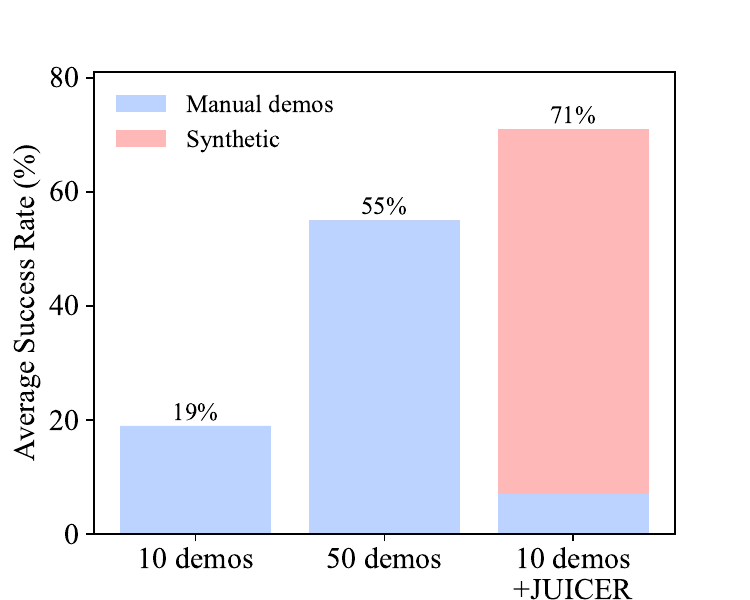}
        \vspace{-4pt}
        \captionof{figure}{\small Learning \oneleg{} from 10 human demos.}
        \label{fig:bootstrap-success}
    \end{minipage}
    \hfill
    \vspace{-20pt}
\end{figure*}
\fi

\subsection{Numerical Results: Success rates}

\autoref{fig:main-results} shows success rates across tasks and methods.

\subsubsection{MLP Baseline and Diffusion Policy}

The MLP policy without action chunking (MLP-NC) could not complete any tasks. With action-chunking (MLP-C), however, the MLP policy drastically improves, achieving average success rates of 40\%, 9\%, 3\%, and 4\% for \oneleg{}, \roundtable{}, \lamp{}, and \squaretable{}, respectively.
The performance difference between predicting the next action $T_p=1$ versus predicting a chunk $T_p=32$ further corroborates that inducing long-term action consistency is highly beneficial for IL~\cite{chi_diffusion_2023,zhao_rss23_aloha}.
We also observe the diffusion policy (DP-BC) to improve performance beyond the MLP. 
The relative performance improvement is greater for the tasks with longer horizons, possibly due to a stronger ability to stay in-distribution\cite{block2024provable}, somewhat mitigating compounding errors.

\subsubsection{Proprioceptive State Noise Injection}

The results from injecting noise into the robot state~\cite{ke_grasping_2021} are mixed: State noising reduces performance for \roundtable{}, makes little difference for \lamp{}, but improves performance for \oneleg{} and \squaretable{} over the standard BC model.
We hypothesize that this reflects the different relative utility of proprioceptive state information depending on the task. For instance, the proprioceptive state is less useful than the egocentric view during insertion in \squaretable{}, whereas for \roundtable{}, the proprioceptive state is likely more helpful because the wrist camera image often becomes occluded by the larger grasped part during insertion. It appears that noise injection harms performance when reliable robot states are critical, but in other cases, it likely offers a beneficial form of regularization, i.e., as discussed in~\cite{hsu_vision-based_2022}.

\subsubsection{Trajectory Augmentation (TA)}

Adding synthetic data around the bottleneck states to the original dataset of 50 teleoperation demonstrations improves performance, especially for \roundtable{} and \oneleg{}. 
We believe that TA provides larger performance gains on these tasks because the primary bottleneck in \roundtable{} and \oneleg{} involves precise insertion. In contrast, the main challenge in \lamp{} is to grasp a dynamic rolling bulb, and our augmentation procedure is currently less suited to overcoming challenges with dynamic objects. For \squaretable{}, the task comprises a long sequence of sub-tasks, many of which are not insertions and thus currently not addressed by the augmentation. Thus, only focusing on bottleneck states creates a more delicate balance between increasing support for some phases and not washing out other phases.

We believe that TA provides larger performance gains on these tasks because the primary bottleneck in \roundtable{} and \oneleg{} involves precise insertion. In contrast, the main challenge in \lamp{} is to grasp a bulb that can roll on the table, introducing a quasi-dynamic element not present in the other tasks. Our current augmentation procedure is less suited to addressing this rolling behavior, as it focuses on static configurations around insertion points. For \squaretable{}, the task comprises a long sequence of sub-tasks, many of which are not insertions and thus currently not addressed by the augmentation. Consequently, focusing solely on bottleneck states creates a more delicate balance between increasing support for some phases and not diluting the importance of others.

\subsubsection{Collect-and-Infer (CI)}

We see a large improvement when training on a mix of 50 teleop demos and 50 successful rollouts collected from arbitrary model evaluations. 
The improvement is most marked for the \lamp{} task, where the success rate increases $3\times$.
This is attributable to the dynamic bulb part that rolls on the table and reaches different table regions depending on how the robot and other nearby parts interact with it. By training on successful rollouts, the policy observes a wider distribution of initial bulb positions. 
We also see a doubling of performance for \squaretable{}, which is likely attributable to the long-horizon and multi-faceted nature of the task, which requires more uniform coverage of the state space throughout the episode.

\subsubsection{Traj. Aug. \& Collect-and-Infer (TA \& CI)}

We hypothesize that synthetic augmentations and rollout trajectories are mutually beneficial, as more augmentations can be added without washing out other phases.
Indeed, the results in \autoref{fig:main-results} show that combining TA \& CI leads to the overall best average performance across all tasks, achieving averages of 74\%, 32\%, 28\%, and 15\% on \oneleg{}, \roundtable{}, \lamp{}, and \squaretable{}, respectively.

\subsubsection{Multitask Training}

% When training on a mix of 50 demonstrations from each of \roundtable{}, \lamp{}, and \squaretable{}, we observe an improvement in performance on each of the individual tasks. % with the same model. 
For multitask training, we find that a \emph{single} model trained on a mix of 50 demonstrations from each of \roundtable{}, \lamp{}, and \squaretable{} outperforms all baseline diffusion policy (DP-BC) models trained individually on each respective task.
The ability to improve performance from 18\% to 25\% for \roundtable{} and 6\% to 12\% for \lamp{} merely by mixing data from different tasks suggests that the model implicitly learns certain skills shared between the tasks.

\subsubsection{Further Minimizing Demonstration Budget}
\label{sec:minimize-demo-budget}
Finally, we perform one full \methodname{} iteration (i.e., collecting demonstrations, augmenting, and performing collect-infer) starting from just 10 demonstrations of the \oneleg{} task and compare it with the performance obtained from 50 human-collected demonstrations.
\autoref{fig:bootstrap-success} shows that training the baseline diffusion policy on only 10 demonstrations for the \oneleg{} task yields a success rate of 19\%, compared to 59\% with 50 demonstrations.
However, models trained on a combined set of 10 human demonstrations, 90 successful rollouts, and 100 augmentations achieve, on average, 71\% success, a $>10\%$ improvement beyond collecting 50 demos by hand.

\section{Related Work}

\subsection{Improving Robustness of Behavior Cloning}
There have been many proposed techniques for perturbing states and obtaining corrective actions to expand dataset support.
In~\cite{ke_grasping_2021}, system states are randomly corrupted, and dataset actions are re-used as corrective actions.
\cite{ke_ccil_2023} generalizes this by learning local dynamics models that support synthesizing recovery actions. % that recover from deviated states.
Each system relies on a fully observed state space and does not readily apply when operating with image observations and unknown object poses. 

The method in~\cite{zhou2023nerf} trains a Neural Radiance Field to synthesize wrist-camera views that would result from perturbed end-effector states. Perturbations are then used to construct corrective action labels. However, \cite{zhou2023nerf} operates with static scenes and assumes an image masking ability that is more challenging for a policy that must insert an unknown grasped object.
Other strategies may involve augmented loss terms~\cite{pfrommer_tasil_2023} or requiring experts to directly demonstrate corrective actions~\cite{laskey_dart_2017, brandfonbrener_visual_2022}. Each of these requires assumptions and complexities that our system does not incur.

\subsection{Diffusion for Decision-Making}
Diffusion models~\cite{ho_denoising_2020,song2020score,sohl2015deep_diffusion} have been shown to provide significant performance improvements in a variety of sequential decision-making and control domains~\cite{ajay_is_2023,chi_diffusion_2023,janner_planning_2022, pearce_imitating_2023,reuss2023goal_diffusion_policy}.
There are mechanisms for predicting actions at high frequencies with diffusion~\cite{chi_diffusion_2023}, and such models have supported the development of theoretical performance guarantees for BC~\cite{block_provable_2023}. 
Our results similarly show an expressive diffusion policy architecture benefits performance in the challenging setting of multi-step assembly from images.

\subsection{Behavior Learning for Insertion and Assembly}
Much past work has studied the related problems of assembly~\cite{tang2023industreal, yash_factory2022_rss, thomas2018learning, zhang2022learning_assembly}, insertion~\cite{beltran2020variable, luo2020dynamic, schoettler2020meta, davchev2022residual}, and kitting~\cite{li2022scene, devgon2021kit}.
InsertionNet~\cite{spector_insertionnet_2021, spector_insertionnet_2022} reverses random actions taken from a manually demonstrated ``inserted'' state to supervise robust insertion behaviors.
Form2Fit~\cite{zakka2020form2fit} uses self-supervised disassembly sequences to learn pick-and-place affordance models for kitting tasks.
Assemble-them-all~\cite{tian2022assemble_them_all} and ASAP~\cite{tian2023asap} also leverage disassembly to guide the search for a sequence and motion that achieves complex multi-part assembly. 
Our work generalizes the same principle for data augmentation around bottleneck transitions while operating on continuous low-level actions and training from complete demonstration trajectories. % 

\section{Limitations and Conclusion}
\paragraph*{Limitations} While simulation offers our pipeline many benefits, we have not yet adapted our approach to the real world. One potential way forward is to directly transfer our learned policies using domain randomization. We are also interested in real-world augmentation/collect-and-infer strategies, which may require additional components for success classification and resets. We also cannot yet generalize to arbitrary initial part pose distributions. However, there is potential of combining collect-and-infer with a curriculum that gradually increases the initial state distribution.

\paragraph*{Conclusion}
This paper proposes a mechanism for training a robot to perform precise, multi-step assembly tasks in simulation using a small demonstration data budget. We develop a pipeline combining expressive policy architectures with tools for synthetic dataset expansion and corrective action generation to overcome challenges with limited dataset size, covariate shift, and task complexity. Our results illustrate that the pipeline can use a modest number of demonstrations to acquire policies for constructing multiple complex assemblies that require precise manipulation behaviors executed over long horizons.


%%%%%%%%%%%%%%%%%%%%%%%%%%%%%%%%%%%%%%%%%%%%%%%%%%%%%%%%%%%%%%%%%%%%%%%%%%%%%%%%

\section*{ACKNOWLEDGMENTS}

This work was partly supported by the Sony Research Award and the US Government. The computations in this paper were run on the FASRC cluster, supported by the FAS Division of Science Research Computing Group at Harvard University, and on the MIT Supercloud\cite{reuther_interactive_2018}.

\textbf{Contributions}
\textbf{LA} led the project, implementation, and experimental evaluation, and contributed to ideation and writing.
\textbf{AS} conceived the initial project idea, advised on project goals, contributed to the  implementation of some parts of the system and writing.
\textbf{IS} actively contributed in research discussions and provided feedback on the paper and writing.
\textbf{PA} advised the project and contributed to its development, experimental design, and positioning.

%%%%%%%%%%%%%%%%%%%%%%%%%%%%%%%%%%%%%%%%%%%%%%%%%%%%%%%%%%%%%%%%%%%%%%%%%%%%%%%%

\bibliographystyle{IEEEtran}
\bibliography{references}

%%%%%%%%%%%%%%%%%%%%%%%%%%%%%%%%%%%%%%%%%%%%%%%%%%%%%%%%%%%%%%%%%%%%%%%%%%%%%%%%

\clearpage
% \fi
\onecolumn % Switch to single-column layout
\section*{Appendix}

\subsection{Implementation Details}

\subsubsection{Training Hyperparameters}
\label{app:hyperparameters}

We present a comprehensive set of hyperparameters used for training. Hyperparameters shared for all models are shown in \autoref{tab:training-hyperparameters}, hyperparameters specific to the diffusion models in \autoref{tab:diffusion-model-hyperparameters}, and hyperparameters specific to the MLP baseline in \autoref{tab:mlp-baseline-hyperparameters}.

\begin{table}[H]
\centering
\caption{Training Hyperparameters Shared for All Models}
\label{tab:training-hyperparameters}
\begin{tabular}{p{0.3\linewidth}p{0.2\linewidth}}
\toprule
Parameter & Value \\
\midrule
Control Mode & Position \\
Action Space Dimension & 10 \\
Observation Space Dimension & 16 \\
Orientation Representation & 6D \\
Max LR & $10^{-4}$ \\
LR Scheduler & Cosine \\
Weight Decay & $10^{-6}$ \\
Warmup steps & 500 \\
Batch Size & 256 \\
Max Epochs & 400 \\
Steps per Epoch & 400 \\
Image Size Input & $2 \times 320 \times 240 \times 3$ \\
Image Size Encoder & $2 \times 224 \times 224 \times 3$ \\
Vision Encoder Model & ResNet18 \\
Encoder Weights & ImageNet 1k \\
Encoder Parameters & $2 \times 11$ million \\
Runs per Condition & 5 \\
Encoder Feature Projection Dim & 128 \\
\bottomrule
\end{tabular}
\end{table}

\begin{table}[H]
\centering
\caption{Diffusion Model Hyperparameters}
\label{tab:diffusion-model-hyperparameters}
\begin{tabular}{p{0.3\linewidth}p{0.2\linewidth}}
\toprule
Parameter & Value \\
\midrule
U-Net Down dims & [256, 512, 1024] \\
U-Net Parameters & 69 million \\
Policy total parameters & 91 million \\
Observation Horizon $T_o$ & 1 \\
Prediction Horizon $T_p$ & 32 \\
Action Horizon $T_a$ & 8 \\
DDPM Training Steps & 100 \\
DDIM Inference Steps & 8 \\
\bottomrule
\end{tabular}
\end{table}

\begin{table}[H]
\centering
\caption{MLP Baseline Hyperparameters}
\label{tab:mlp-baseline-hyperparameters}
\begin{tabular}{p{0.3\linewidth}p{0.2\linewidth}}
\toprule
Parameter & Value \\
\midrule
Residual Blocks & 5 \\
Residual Block Width & 1024 \\
Parameters & 10 million \\
Policy total parameters & 32 million \\
Observation Horizon $T_o$ & 1 \\
Prediction Horizon $T_p$ (Chunked) & 32 \\
Action Horizon $T_a$ (Chunked) & 8 \\
Prediction Horizon $T_p$ (No Chunking) & 1 \\
Action Horizon $T_a$ (No Chunking) & 1 \\
\bottomrule
\end{tabular}
\end{table}

\subsubsection{Normalization}

All 10 and 16 dimensions of the action and proprioceptive state, respectively, were independently normalized to lie in the range [-1, 1]. The normalization limits were calculated across all the demonstration data across all 4 tasks to ensure consistent action and state spaces across tasks. This follows the normalization used in, e.g., \cite{reuss_goal-conditioned_2023,chi_diffusion_2023} and is also the generally accepted normalization used for diffusion models. \cite{reuss_goal-conditioned_2023} standardized the input to have mean 0 and standard deviation 1 instead of min-max scaling to [0,1], which is something we did not test in our experiments.

\subsubsection{Rotation Representation}
\label{app:rotation-representation}

We represent all orientations and rotations with the 6D representation, both for the predicted action and proprioceptive end-effector pose orientation \cite{zhou2019continuity,levinson2020analysis}. The end-effector rotation angular velocity was still encoded as roll, pitch, and yaw values. This representation of rotations contains redundant dimensions but is continuous in that small changes in orientation also lead to small changes in the values in the representation, which is not generally the case for Euler angles and quaternions, which can enable easier learning.

\subsubsection{Image Augmentation}

We apply image augmentation to both images during training. We apply random cropping only to the front camera view. In addition, we apply color jitter with hue, contrast, brightness, and saturation set to 0.3 and Gaussian blur with a kernel of size 5 and sigma between 0.1 and 5 to both camera views.

At inference time, we statically center-crop the front camera image from $320\times 240$ to $224\times 224$ and resize the wrist camera view with the same dimensions. For both the random crop and center crop, we resize the image to $280\times 240$ to ensure we are not moving the image around so much that essential parts of the scene are cropped out.

The specific choice of the above values was made by eye to find a level that was sufficiently adversarial while still keeping all essential features discernible. We show examples of augmentations below.

\begin{figure}[H]
    \centering
    \includegraphics[width=0.49\textwidth]{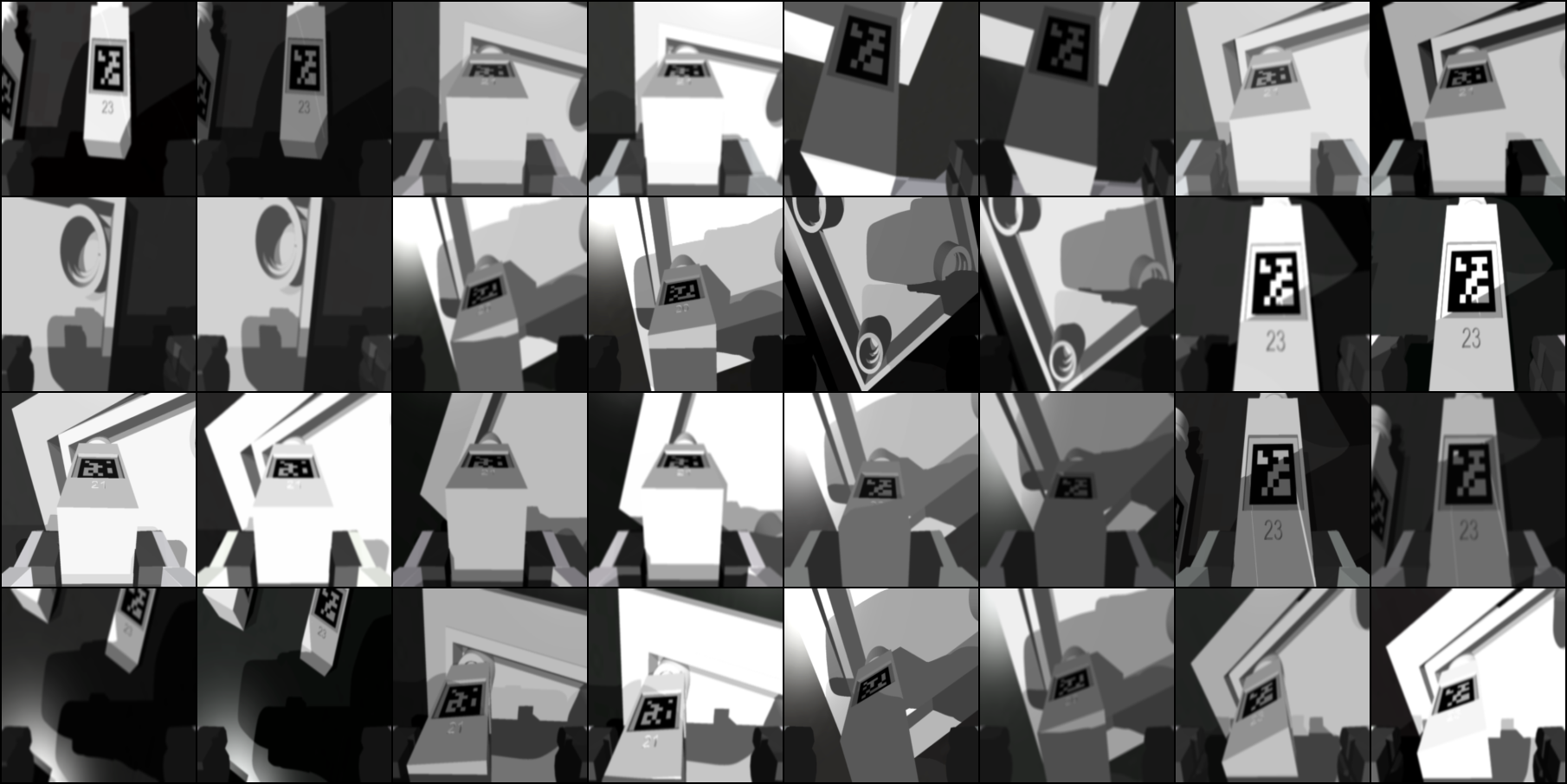}
    \hfill
    \includegraphics[width=0.49\textwidth]{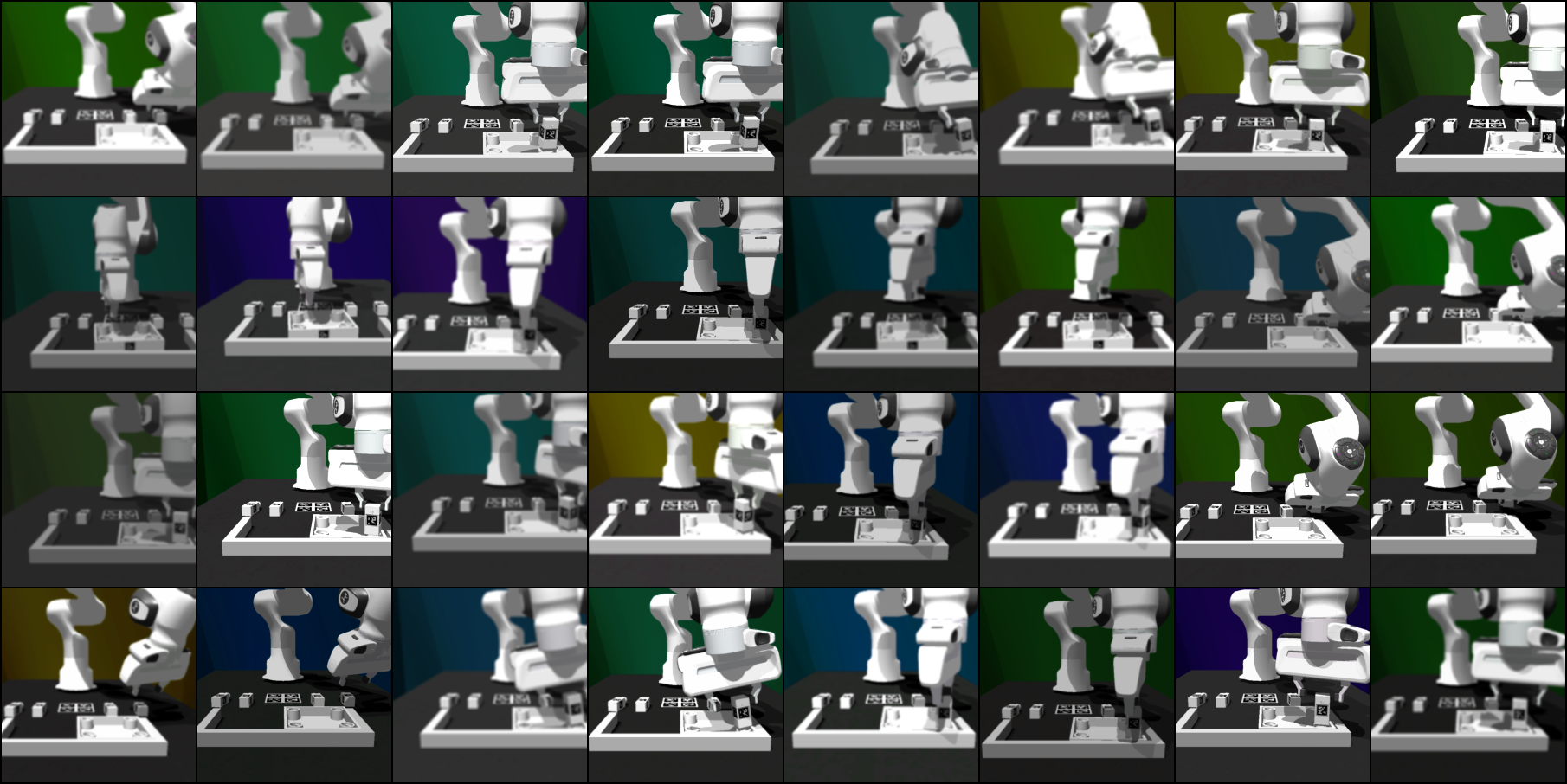}
    \caption{\textbf{Left:} Examples of augmentations of the wrist camera view, consisting of color jitter and Gaussian blur. \textbf{Right:} Examples of augmentations for the front view also consist of color jitter and Gaussian blur augmentations, as well as random cropping.}
    \label{fig:image-aug-example-side-by-side}
\end{figure}

\subsection{Data Collection, Training, and Evaluation Details}

The full pipeline for training the final models with \methodname{} involved several steps and different code files. In the below list, we give a brief summary of the steps on a high level and what files the relevant code and command-line arguments are found in. Please follow the installation instructions in the code repository (\href{https://github.com/ankile/imitation-juicer}{github.com/ankile/imitation-juicer}) to ensure all required packages are installed.

\subsubsection{Teleoperation Demonstration and Annotation Effort}

In \autoref{tab:demo-collection-times}, we present the approximate time the teleoperator spent collecting and annotating 50 demos for each of the 4 tasks in this work. The labeling time includes all time spent on the labeling task, i.e., all idle time resulting from waiting for files to load and write and any mistakes that were undone and redone. Particularly, the loading of the data took a meaningful amount of time, and the whole process could be made significantly faster by optimizing the read-write speeds (which we did not do).

\begin{table}[H]
\centering
\caption{Approximate Demo Collection and Annotation Times in Minutes}
\label{tab:demo-collection-times}
\begin{tabular}{lccc}
\toprule
Task & Critical States & Collect 50 Demos (min) & Annotate 50 Demos (min) \\
\midrule
\oneleg{} & 1 & $\sim$70 & $\sim$10 \\
\roundtable{} & 3 & $\sim$85 & $\sim$20 \\
\lamp{} & 5 & $\sim$95 & $\sim$30 \\
\squaretable{} & 4 & $\sim$210 & $\sim$30 \\
\bottomrule
\end{tabular}
\end{table}

\subsubsection{Simulator Teleoperation Assistive Improvements}

\begin{figure}[H]
    \centering
    \includegraphics[width=0.5\textwidth]{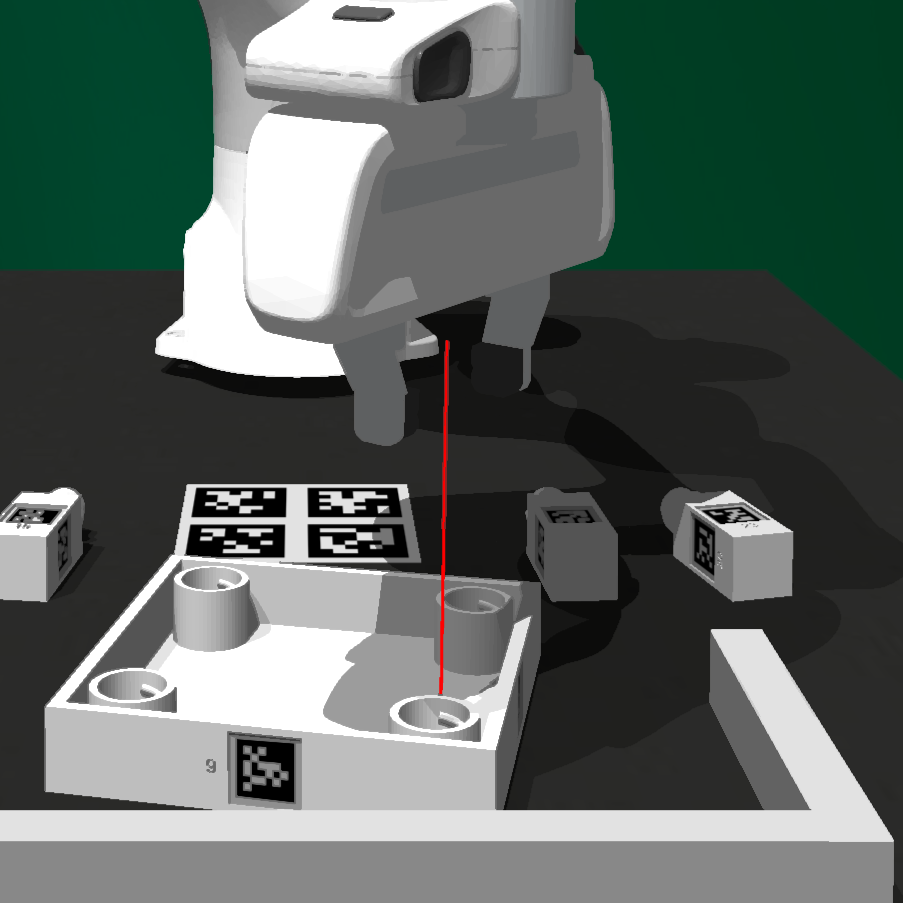}
    \caption{The red light is added as a visual aid for the teleoperator to help alleviate the difficulties introduced by the lack of depth perception. The red line is only visible to the operator and is not rendered to the image observations that are stored while teleoperating. In the image, in the absence of the line, it is quite hard to tell if the end-effector is placed right above the tabletop or not. With the red line, however, it becomes much more obvious.}
    \label{fig:ee-laser}
\end{figure}

One crucial difficulty with collecting data in a simulator is the lack of depth perception. This impediment is, on the surface, very limiting. However, two things helped alleviate the difficulties. First is our experience that the brain is quite adaptable and very readily learned to rely on cues other than depth to infer the relative positions of objects in the depth direction (presumably shadows and relative sizes). Second, we added a ``laser light'' that protrudes out from the end-effector that more explicitly indicates the position and orientation of the end-effector in the scene, as shown in \autoref{fig:ee-laser}.

\subsubsection{Dataset Sizes Across Tasks and Methods}
\label{app:dataset-sizes}

This section details more about what mix of data from the different sources the different models across normal behavior cloning (BC), trajectory argumentation (TA), Collect-Infer (CI), and a mix of all of them for the models for which we report results in the main results figure and \autoref{tab:main-results}.

\begin{table}[H]
\centering
\caption{Dataset Size for Task and Method}
\label{tab:dataset-summary}
\sisetup{group-separator={,}, group-minimum-digits={3}, table-format=3.0}
\begin{tabular}{
  l 
  l 
  S[table-format=2.0] 
  S[table-format=3.0k] 
  S[table-format=2.0] 
  S[table-format=3.0k] 
  S[table-format=3.0] 
  S[table-format=3.0k] 
  S[table-format=3.0k]
}
\toprule
& & \multicolumn{2}{c}{Teleop} & \multicolumn{2}{c}{Rollout} & \multicolumn{2}{c}{Augmentation} & \\
\cmidrule(lr){3-4} \cmidrule(lr){5-6} \cmidrule(lr){7-8}
{Task} & {Method} & {Demos} & {Timesteps} & {Demos} & {Timesteps} & {Demos} & {Timesteps} & {Total Timesteps} \\
\midrule
one\_leg & BC & 50 & 29k & {--} & {--} & {--} & {--} & 29k \\
& Traj. Aug & 50 & 29k & {--} & {--} & 50 & 3k & 32k \\
& Col. Inf. & 50 & 29k & 50 & 27k & {--} & {--} & 56k \\
& TA + CI & 50 & 29k & 50 & 27k & 150 & 8k & 64k \\
\addlinespace
round\_table & BC & 50 & 47k & {--} & {--} & {--} & {--} & 47k \\
& Traj. Aug & 50 & 47k & {--} & {--} & 500 & 12k & 59k \\
& Col. Inf. & 50 & 47k & 50 & 55k & {--} & {--} & 102k \\
& TA + CI & 50 & 47k & 50 & 55k & 700 & 17k & 119k \\
\addlinespace
lamp & BC & 50 & 42k & {--} & {--} & {--} & {--} & 42k \\
& Traj. Aug & 50 & 42k & {--} & {--} & 100 & 5k & 47k \\
& Col. Inf. & 50 & 42k & 50 & 45k & {--} & {--} & 87k \\
& TA + CI & 50 & 42k & 50 & 45k & 200 & 10k & 97k \\
\addlinespace
square\_table & BC & 50 & 130k & {--} & {--} & {--} & {--} & 130k \\
& Traj. Aug & 50 & 130k & {--} & {--} & 150 & 8k & 138k \\
& Col. Inf. & 50 & 130k & 50 & 126k & {--} & {--} & 256k \\
& TA + CI & 50 & 130k & 50 & 126k & 400 & 22k & 278k \\
\bottomrule
\end{tabular}
\end{table}

\subsection{Denoising Diffusion Probabilistic Models}
\label{app:denoising-diffusion}

Denoising Diffusion Probabilistic Models (DDPM) \cite{ho_denoising_2020,song2020score,sohl2015deep_diffusion} are a powerful class of generative models used to learn and sample from high-dimensional distributions. This appendix provides a detailed explanation of the DDPM framework, including its mathematical foundations and key concepts.

\paragraph{Overview}

DDPMs work on the principle of gradually adding noise to data and then learning to reverse this process. This approach allows the model to generate new samples by starting from pure noise and progressively refining it into coherent data.

\paragraph{The Forward Process}

The forward process, also known as the diffusion process, is a fixed Markov chain that gradually adds Gaussian noise to the data. For a data point $x_0$, we define a sequence of latent variables $x_1, ..., x_{K-1}, x_K$, where $K$ is the total number of steps in the process.

The forward process is defined as:

\begin{equation}
    q(x_{k+1}|x_{k}) = \mathcal{N}(x_{k+1}; \sqrt{\alpha_k}x_k, (1-\alpha_k)\mathbf{I})
\end{equation}

Here, $\alpha_k$ is a noise schedule that determines how much noise is added at each step. The process is designed such that $x_K$ becomes pure Gaussian noise: $x_K \sim \mathcal{N}(\mathbf{0}, \mathbf{I})$.

\paragraph{The Reverse Process}

The reverse process, which is learned during training, aims to gradually denoise the data, starting from pure noise. It is modeled as:

\begin{equation}
    p_\theta(x_{k-1}|x_k) = \mathcal{N}(x_{k-1}|\mu_\theta(x_k,k), \Sigma_k)
\end{equation}

where $\theta$ represents the parameters of the model, typically a neural network.

\paragraph{Training Objective}

To train the model, we use a ``noise prediction'' surrogate loss. This approach is made possible by the ability to sample from any intermediate step of the forward process directly:

\begin{equation}
    q(x_k|x_0) = \mathcal{N}(\sqrt{\Bar{\alpha}_k} x_0, (1-\Bar{\alpha}_k)\mathbf{I})
\end{equation}

where $\Bar{\alpha}_k = \prod_{t=1}^k \alpha_t$.

The training objective is then formulated as:

\begin{equation}
    \mathcal{L}(\theta) = \mathbb{E}_{\substack{k\sim [K]\\x_0\sim q\\ \epsilon\sim\mathcal{N}(0, I)}} \left[ \parallel \epsilon - \epsilon_\theta (\sqrt{\Bar{\alpha}_k}\mathbf{x}_0 + \sqrt{1-\Bar{\alpha}_k}\epsilon, k)\parallel^2 \right]
\end{equation}

Here, $\epsilon_\theta$ is a neural network that predicts the noise added to the input.

\paragraph{Sampling Process}

After training, we can generate new samples by starting from Gaussian noise $x_K$ and iteratively applying the learned reverse process:

\begin{equation}
    x_{k-1} = \frac{1}{\sqrt{\alpha_k}}\left(x_k - \frac{1-\alpha_k}{\sqrt{1-\Bar{\alpha_k}}}\epsilon_\theta(x_k, k)\right) + \sigma_k \mathbf{z}
\end{equation}

where $\mathbf{z} \sim \mathcal{N}(0, \mathbf{I})$ and $\sigma_k$ is a small noise term added for stochasticity.

\paragraph{Improved Sampling}

The DDIM sampler \cite{song_denoising_2022,lu2022dpm} provides a method for deterministic sampling with fewer steps, improving the efficiency of the generation process while maintaining quality. This is achieved by modifying the noise prediction process and allowing for a more flexible trajectory between the noise and the final sample.

\subsection{Results in Tabular Form}

In \autoref{tab:main-results}, we present the same results as in the main results table, but in tabular form with the exact numbers.

\begin{table}[H]
    \centering
    \caption{Main Results Table}
    \label{tab:main-results}
    \begin{tabular}{lccccccccccccc}
    \toprule
    Task & \multicolumn{2}{c}{One leg} & \multicolumn{2}{c}{Round table} & \multicolumn{2}{c}{Lamp} & \multicolumn{2}{c}{Square table} \\
    \cmidrule(lr){2-3} \cmidrule(lr){4-5} \cmidrule(lr){6-7} \cmidrule(lr){8-9}
        Method     & Avg & Max & Avg & Max & Avg & Max & Avg & Max \\
    \midrule
    MLP-NC      & 0   & 0   & 0   & 0   & 0 & 0 & 0 & 0 \\
    MLP-C       & 40  & 52  & 9   & 15  & 3   & 4   & 4   & 9   \\
    DP-BC          & 59  & 68  & 18  & 21  & 6   & 7   & 6   & 9   \\
    State noise~\cite{ke_grasping_2021} & 64  & 70  & 12  & 19  & 6   & 11  & 13  & 16  \\
    \textbf{Traj. Aug.}     & 64  & 65  & 28  & 31  & 8   & 12  & 9   & 17  \\
    \textbf{Col.-Inf.}      & \textbf{75}  & 79  & 24  & 27  & 18  & \textbf{29}  & 12  & \textbf{20}  \\
    \textbf{TA \& CI}       & \textbf{74}  & \textbf{85}  & \textbf{32}  & \textbf{35}  & \textbf{28}  & \textbf{30}  & \textbf{15}  & 17  \\
    \textbf{Multi-task}     & 58  & 67  & 25  & \textbf{34}  & 12  & 18  & 7   & 10  \\
    \bottomrule
    \end{tabular}
\end{table}

\subsection{Further Analysis}

\subsubsection{Data Efficiency}

\begin{figure}[H]
\centering
\includegraphics[width=0.75\linewidth]{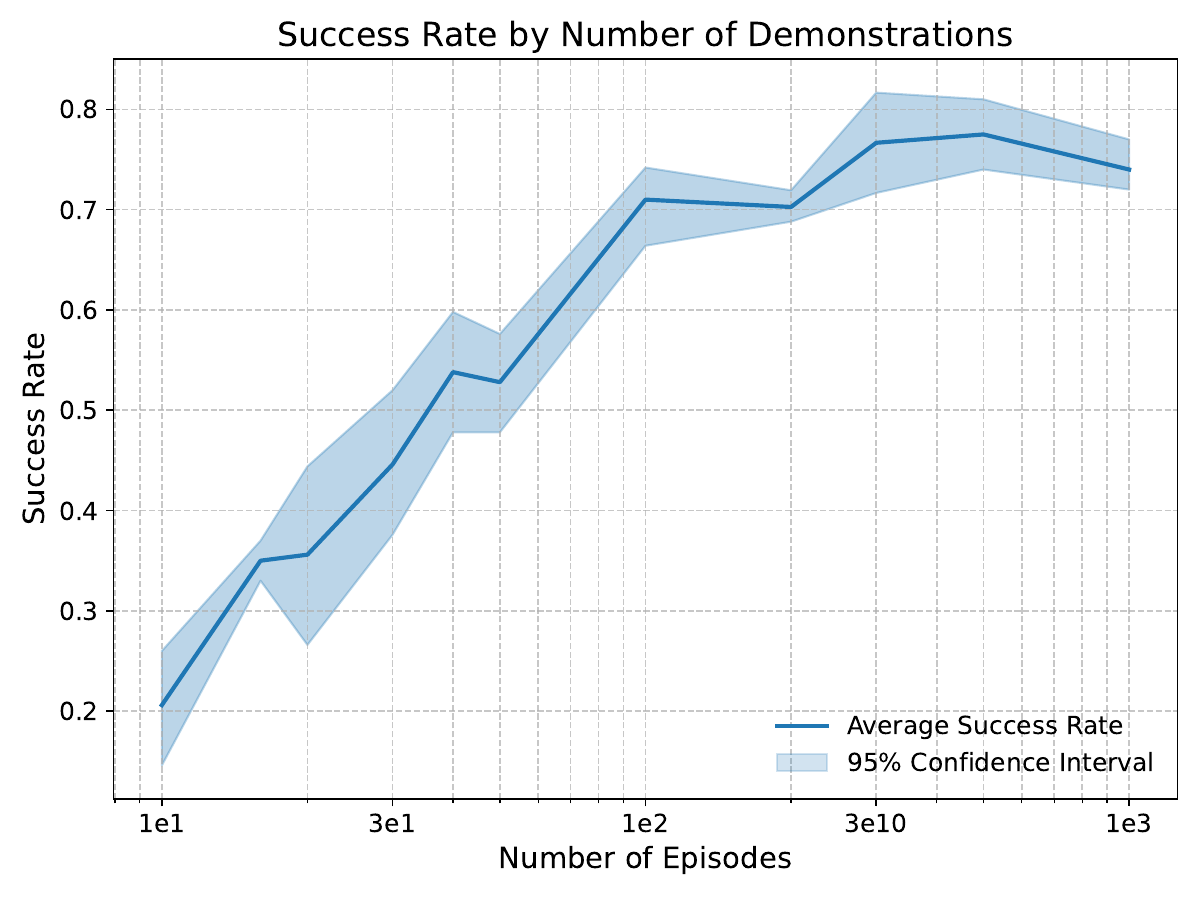}
\caption{Data efficiency graph for the `one\_leg` task.}
\label{fig:data-efficiency}
\end{figure}

In \autoref{fig:data-efficiency}, we see how the performance for the \oneleg{} task changes with increasing demo dataset size. We calculate success rates for policies trained with [10, 20, 30, 40, 50, 100, 200, 300, 500, 1000] demos. We find the success rate to follow an exponential curve closely up to 1000 demonstrations but seems to saturate at around $\sim$300-500 demonstration trajectories in the training data.

\subsubsection{Number of Denoising Steps in DDIM Sampler}

\begin{figure}[H]
\centering
\begin{subfigure}[t]{0.45\linewidth}
\includegraphics[width=\linewidth]{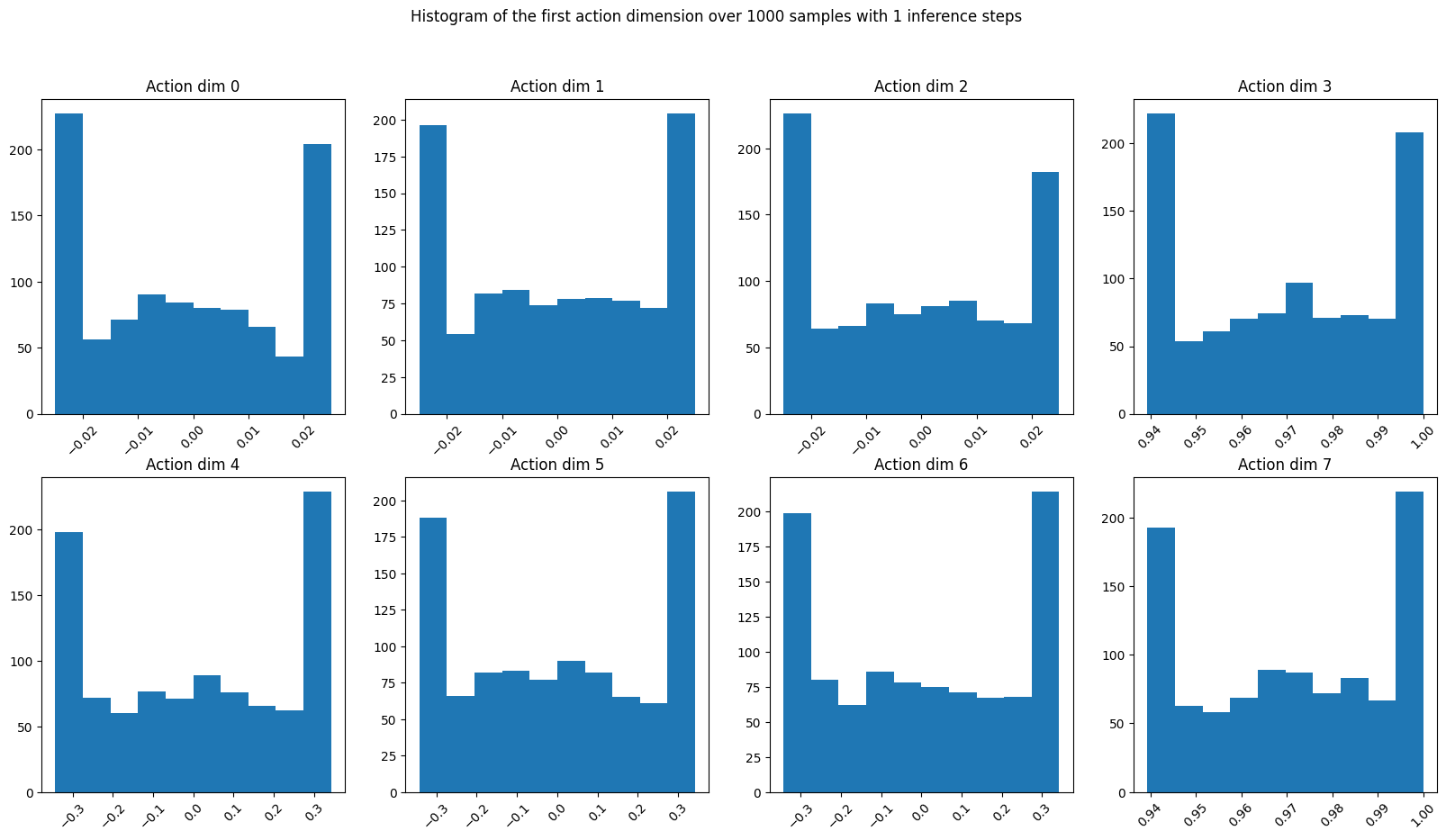}
\caption{Action prediction using 1 DDIM sampling step.}
\end{subfigure}
\quad
\begin{subfigure}[t]{0.45\linewidth}
\includegraphics[width=\linewidth]{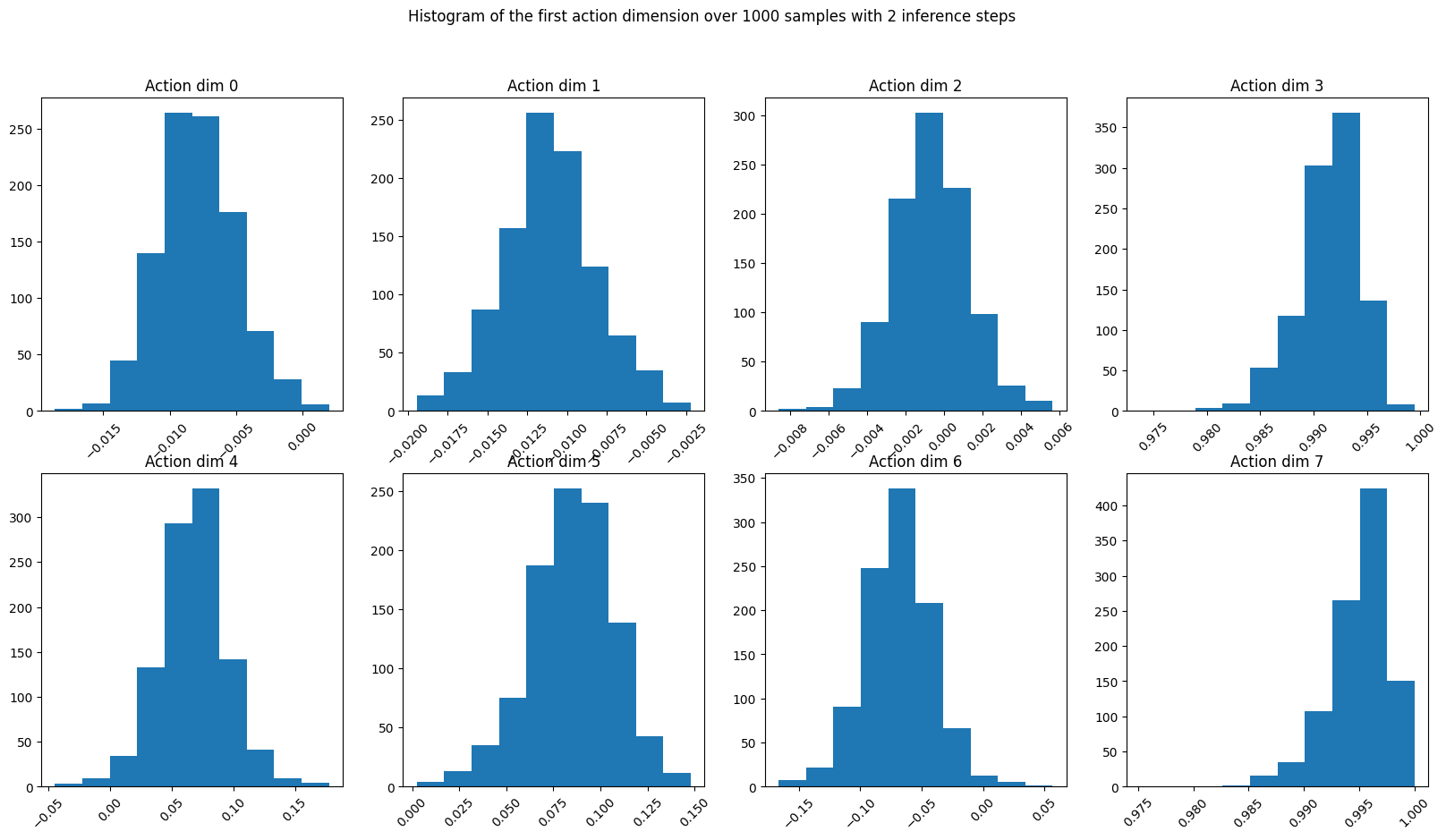}
\caption{Action prediction using 2 DDIM sampling steps. The samples are already getting relatively good.}
\end{subfigure}

\vspace{1em}

\begin{subfigure}[t]{0.45\linewidth}
\includegraphics[width=\linewidth]{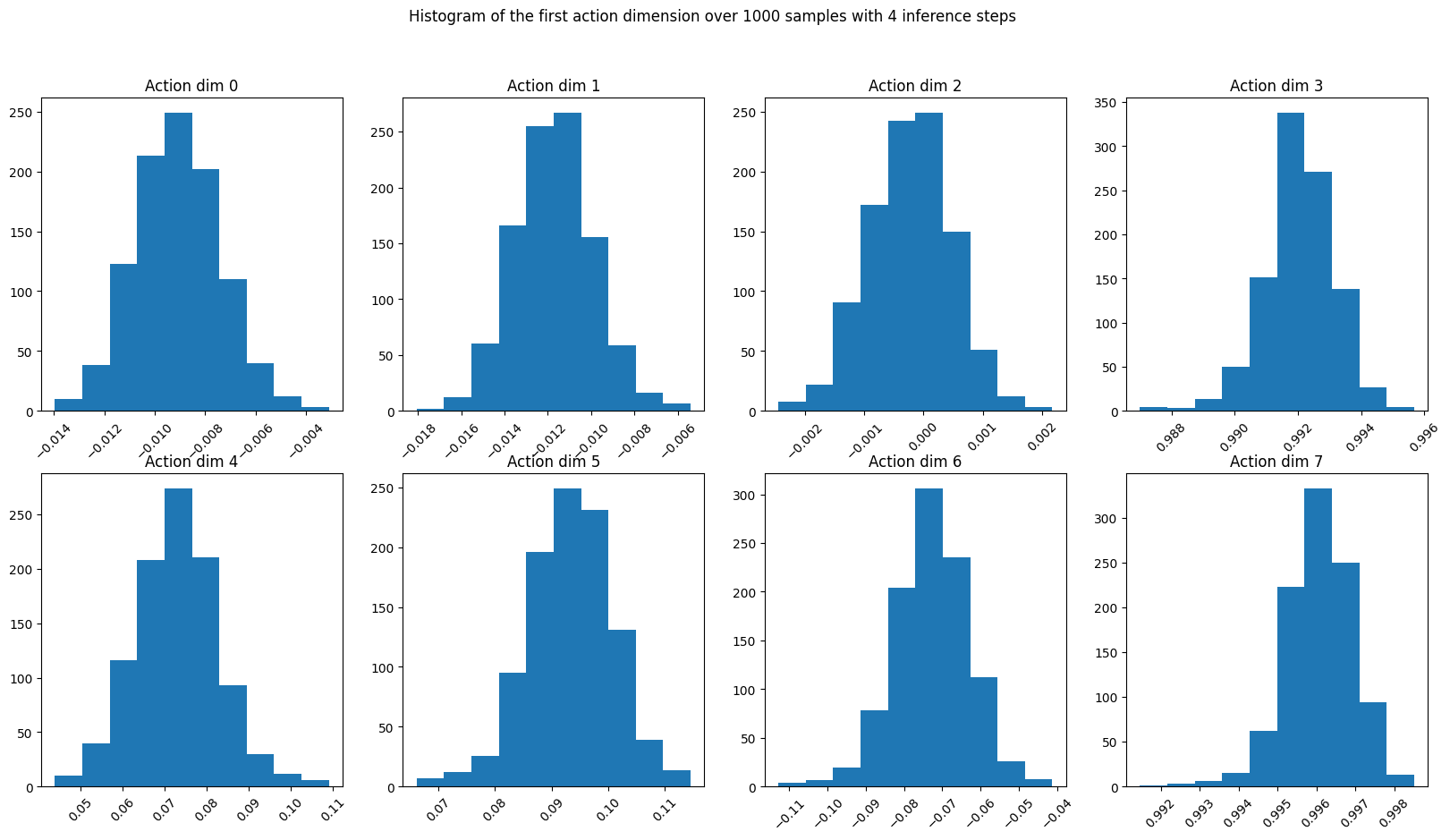}
\caption{Action prediction using 4 DDIM sampling steps. The samples are very hard to distinguish from the predictions using 100 DDIM steps.}
\end{subfigure}
\quad
\begin{subfigure}[t]{0.45\linewidth}
\includegraphics[width=\linewidth]{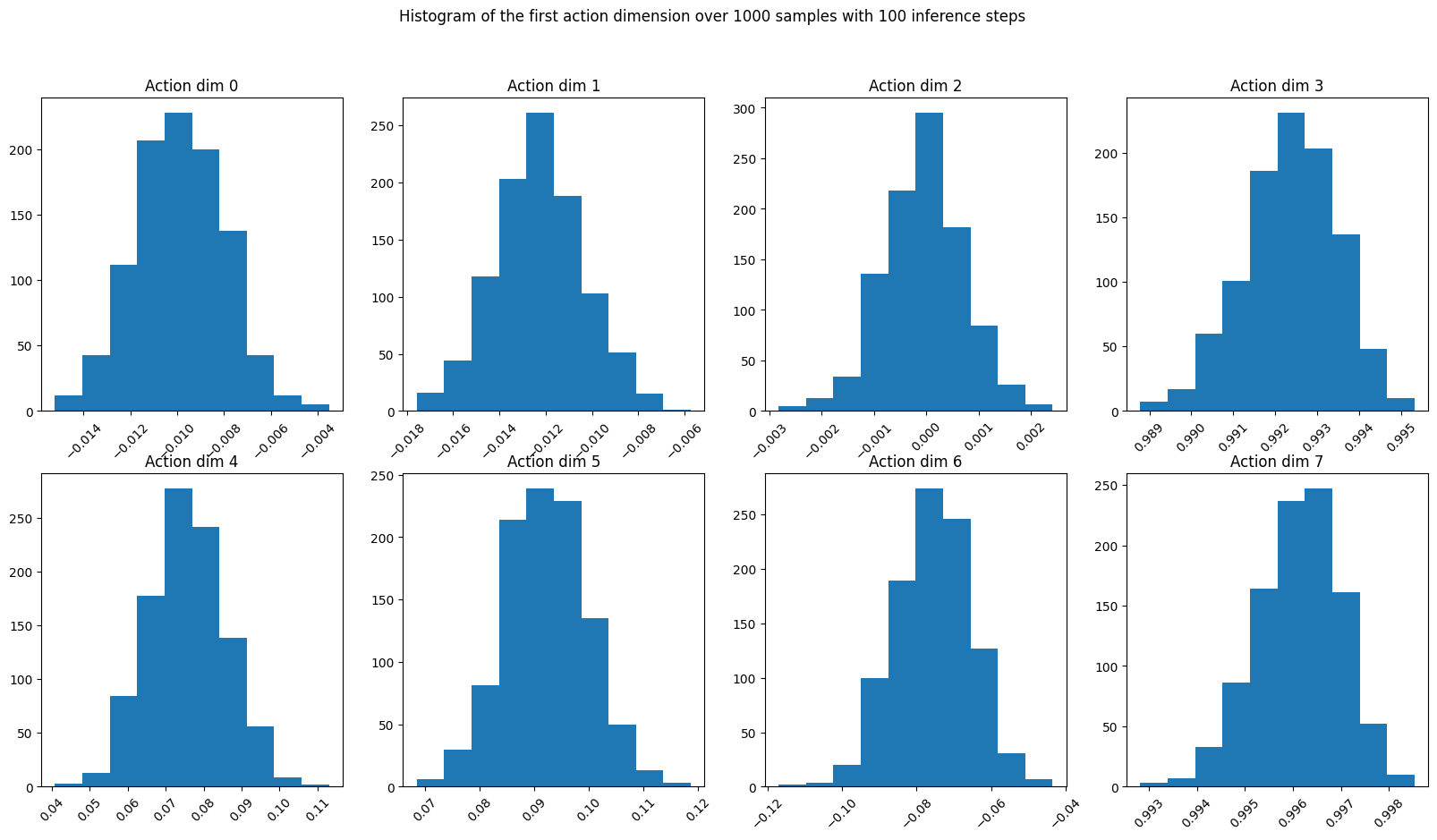}
\caption{Action prediction using 100 DDIM sampling steps.}
\end{subfigure}

\caption{Comparison of samples of predicted actions for the same state with a varying number of DDIM sampling steps. The four plots show the resulting prediction sample histograms for 1000 action samples with 1, 2, 4, and 100 DDIM sampling steps, respectively. Each of the 8 histograms in each of the 4 subplots shows the distribution for a specific dimension of the action space: $[\Delta x, \Delta y, \Delta z, \Delta \text{roll}, \Delta \text{pitch}, \Delta \text{yaw}, \text{gripper-width}]$.
Increasing the sampling steps from 1 to 2 produces a vast difference in the distribution, but further increases do not change the predicted action distributions much.}
\label{fig:action-dist-ddim-steps}

\end{figure}

In \autoref{fig:action-dist-ddim-steps}, we present histograms over samples of actions for a given state observation for different numbers of DDIM inference sampling steps of 1, 2, 4, and 100. In particular, we chose an arbitrary state during the \oneleg{} task (the robot was grasping a leg and close to the point of insertion in this example). We sampled 1000 random Gaussian noise vectors that we fully denoised using the DDIM sampler with the given number of denoising steps.

In each of the four figures in \autoref{fig:action-dist-ddim-steps}, we show the distribution as a histogram for each action independently, here as delta actions for the 3D position and rotation as roll, pitch, and yaw. The 8th action distribution is the gripper action. These representations are used in this analysis because it is easier to intuitively interpret than the absolute position actions and 6D rotation representations.

We observe that the distributions converge quickly to the ``final" distribution with the number of denoising steps. After 4 steps, it is already hard to distinguish the resulting distribution from the one using 100 steps.

\subsubsection{Ratio of Synthetic to Demonstration Data}

The optimal number of synthetic trajectory snippets was not a question of more being necessarily better, and we experimented with different ratios. Across our tasks, we typically got the best results with the augmentation timesteps constituting somewhere between 8-20\% of the total number of timesteps in the training data (see \autoref{tab:dataset-summary}).

\subsubsection{Importance of Action-Chunking}

In our experiments with implementing the baseline MLP model, we see a drastic difference in performance between the same architecture trained to predict the next action \(T_p=1\) versus ones trained to predict a chunk \(T_p=32\). In particular, not a single full task completion was observed for any rollout for any of the 5 models trained for any of the 4 tasks. Qualitatively, when observing the rollouts, the behavior is also a lot more erratic and less smooth with a prediction horizon of 1.

\subsubsection{Comparison of Different Vision Encoders}

We found a standard ResNet\cite{he2016deep} from the \texttt{torchvision} package with the \texttt{IMAGENET1k}\cite{deng2009imagenet} pretraining weights to work better than both the same ResNet with no pretraining and the ResNet with the spatial softmax pooling layer from \texttt{Robomimic}\cite{robomimic2021} that was used in \cite{chi_diffusion_2023}. We also see very similar performance when using a Vision Transformer\cite{dosovitskiy2021image} model ViT-Small with Dino V1 weights\cite{caron2021emerging}, a ViT-Base with pertaining weights from pertaining on ImageNet with the MAE\cite{he2021masked} objective and a ResNet50 with weights from pertaining on the Ego4D\cite{grauman2022ego4d} dataset with the VIP\cite{ma_vip_2023} objective. The best-performing model, though, was a ResNet18 with pertaining weights from the R3M\cite{nair_r3m_2022} objective on the Ego4D dataset, achieving an average success rate of 77\% for the \oneleg{} task versus 59\% for the ResNet18 pretrained on ImageNet. Surprisingly, the R3M models using the ResNet 34 and 50 performed slightly worse than the ResNet18.

Results for the \oneleg{} tasks for different vision encoder architectures and pretrained weights are summarized in \autoref{tab:vision-encoder-comparison}.

\begin{table}[H]
\centering
\caption{Comparison of Vision Encoder Success Rates on the \oneleg{} task}
\begin{tabular}{lllcc}
\toprule
Model & Training Objective & Supervision & Pretraining Dataset & Success Rate \\
\midrule
ResNet18\cite{he2016deep} & Classification & Supervised & ImageNet\cite{deng2009imagenet} & 59\% \\
ResNet18 & No Pretraining & --- & --- & 49\% \\
ResNet18 (Spatial Softmax)\cite{finn2016deep} & No Pretraining & --- & --- & 30\% \\
ViT-Small\cite{dosovitskiy2021image} & DINO\cite{caron2021emerging} & Self-supervised & ImageNet & 55\% \\
ViT-Base & MAE\cite{he2021masked} & Self-supervised & ImageNet & 63\% \\
\textbf{ResNet18} & \textbf{R3M\cite{nair_r3m_2022}} & \textbf{Self-supervised} & \textbf{Ego4D\cite{grauman2022ego4d}} & \textbf{77\%} \\
ResNet34 & R3M & Self-supervised & Ego4D & 75\% \\
ResNet50 & R3M & Self-supervised & Ego4D & 73\% \\
ResNet50 & VIP\cite{ma_vip_2023} & Self-supervised & Ego4D & 54\% \\
\bottomrule
\end{tabular}
\label{tab:vision-encoder-comparison}
\end{table}

\subsubsection{Empirical Mode Sampling}

During development, we observed that the model can sample actions in low-density parts of the distribution, which could be actions that take the agent out of the distribution, causing a failure. One interesting finding is that as soon as the model was OOD, it tended to widen its action prediction distributions by approximately 10 times.

In experiments with forcing the model to sample only actions near the mode, we implemented a simple Kernel Density Estimation method for estimating the mode of an empirical distribution over a sample of continuous actions. In our experience, this did not improve success rates for the preliminary experiments on \oneleg{}.

These experiments were carried out before reading the results of \cite{pearce_imitating_2023}. We hypothesize that one of the reasons we did not find KDE mode sampling to help in our case is that our joint action-chunk space is of \(|a|\cdot T_a = 10 \cdot 8=80\) dimensions, which can be a very sparsely populated space, even with 5000 samples. This is subject to further investigations in future work.

\end{document}